\documentclass{article}
\usepackage{amsmath, amssymb, amsfonts}
\usepackage{algorithm}
\usepackage{algpseudocode}
\usepackage{booktabs}
\usepackage{graphicx}
\usepackage{multirow}
\usepackage{geometry}
\usepackage{xcolor}
\usepackage{url}
\usepackage{hyperref}
\usepackage{orcidlink}

\title{\textbf{HAS-VQ}: Hessian-Adaptive Sparse Vector Quantization for \\ High-Fidelity LLM Compression}
\author{
  \textbf{Vladimer Khasia \orcidlink{0009-0002-3320-8142}} \\
  Independent Researcher \\
  \texttt{vladimer.khasia.1@gmail.com}
}
\date{January 11, 2026}

\begin{document}

\maketitle

\begin{abstract}
Post-training quantization is essential for deploying Large Language Models (LLMs) on resource-constrained devices. However, standard integer quantization (e.g., INT4) fundamentally degrades performance by imposing a uniform grid on the heavy-tailed distribution of weight parameters, particularly in smaller-scale models (e.g., $<$2B parameters). We introduce \textbf{HAS-VQ} (Hessian-Adaptive Sparse Vector Quantization), a compression framework that strictly decouples high-sensitivity outliers from the bulk weight distribution using second-order sensitivity analysis. HAS-VQ employs a \textbf{Hessian-Masked Decoupling} strategy to isolate sensitive parameters, followed by robust Vector Quantization (VQ) of the remaining dense body. Crucially, we introduce a \textit{residual sparse feedback} mechanism that corrects quantization errors in the most sensitive dimensions, ensuring exact reconstruction of outliers. We evaluate HAS-VQ on \texttt{SmolLM2-1.7B}, demonstrating two distinct regimes of superiority: (1) \textbf{Pareto Dominance over Integer Baselines:} At 4.23 effective bits-per-parameter (BPP), we achieve a perplexity of 14.23, significantly outperforming the standard INT4 baseline (20.03 PPL at 4.71 BPP). (2) \textbf{High-Fidelity Compression:} Relative to the FP16 baseline, HAS-VQ achieves a $\mathbf{2.3\times}$ reduction in model size (7.03 BPP) while maintaining statistically indistinguishable perplexity (10.12 vs. 10.04), effectively offering a lossless compression alternative for bandwidth-constrained environments.

The code is available at {\url{https://github.com/VladimerKhasia/HASVQ}}
\end{abstract}

\section{Introduction}

The deployment of Large Language Models (LLMs) is fundamentally constrained by the memory bandwidth wall. As model parameters scale, storing weights in half-precision (FP16) becomes prohibitive for edge devices. Post-Training Quantization (PTQ) \cite {pmlr-v119-nagel20a, Frantar2022GPTQAP} has emerged as the standard solution, typically reducing bit-width via integer mapping (e.g., INT4/INT8).

However, uniform integer quantization imposes a rigid structural constraint: it assumes an equidistant discretization of the weight space. This assumption fails for the heavy-tailed, non-Gaussian distributions characteristic of Transformer weights \cite{Dettmers2022GPT3int88M, Vaswani2017AttentionIA}. The degradation is particularly acute in smaller models (1B--3B parameters), which lack the localized redundancy to absorb quantization noise, often resulting in perplexity collapse under standard Round-To-Nearest (RTN) INT4 schemes.

Vector Quantization (VQ) offers a theoretically superior alternative by mapping blocks of weights to learned centroids in a continuous space \cite{Gray1984VectorQ}, thereby aligning the codebook with the true probability density of the weights. Yet, standard VQ algorithms (e.g., K-Means) are unstable in the presence of outliers; a single high-magnitude parameter can skew centroid convergence, degrading the reconstruction of the entire weight block.

To address this, we introduce \textbf{HAS-VQ} (Hessian-Adaptive Sparse Vector Quantization). Our approach rests on the observation that while the majority of weights form a dense, quantizable body, a sparse subset of parameters possesses high Fisher Information and must be preserved with high precision to maintain the optimization trajectory \cite{LeCun1989OptimalBD}.

Our contributions are as follows:
\begin{enumerate}
    \item \textbf{Hessian-Masked Decoupling:} We propose a rigorous decomposition strategy that uses diagonal Hessian sensitivity to orthogonalize outliers from the dense body, ensuring that VQ centroids fit the bulk distribution without skew.
    \item \textbf{Residual Sparse Feedback:} Unlike methods that simply prune or isolate weights, we implement a residual correction mechanism where the sparse component explicitly compensates for the quantization error $\mathcal{E} = \mathbf{W} - \mathcal{Q}(\mathbf{W})$ in the most sensitive dimensions.
    \item \textbf{High-Fidelity Compression:} We empirically demonstrate on \texttt{SmolLM2-1.7B \cite{allal2025smollm2smolgoesbig}} that HAS-VQ offers a $\mathbf{2.3\times}$ reduction in storage compared to FP16 with negligible loss in fidelity (0.08 PPL difference), establishing it as a viable ``near-lossless'' compression format.
    \item \textbf{Efficiency vs. Integer Grids:} We show strict Pareto dominance over standard INT4 baselines, reducing perplexity by $29\%$ while requiring $11\%$ less storage (4.23 BPP vs. 4.71 BPP).
\end{enumerate}

\section{Methodology}

We propose \textbf{HAS-VQ}, a framework designed to minimize the Hessian-weighted quantization noise by explicitly decoupling high-sensitivity outliers from the low-frequency weight distribution.

\subsection{Problem Formulation}
Consider a layer weight matrix $\mathbf{W} \in \mathbb{R}^{d_{out} \times d_{in}}$. Our objective is to find a compressed representation $\widehat{\mathbf{W}}$ that minimizes the expected increase in task loss $\mathcal{L}$. Approximating the loss landscape via a second-order Taylor expansion and assuming a diagonal Fisher Information Matrix (Hessian) $\mathbf{H}$, the objective is to minimize the Hessian-weighted Mean Squared Error:

\begin{equation}
    \min_{\widehat{\mathbf{W}}} \mathbb{E} \left[ (\mathbf{W} - \widehat{\mathbf{W}})^\top \mathbf{H} (\mathbf{W} - \widehat{\mathbf{W}}) \right] \approx \sum_{i,j} \mathbf{H}_{ii} (\mathbf{W}_{ij} - \widehat{\mathbf{W}}_{ij})^2
\end{equation}

\subsection{Algorithm Design}

HAS-VQ executes a four-stage compression pipeline: Sensitivity Analysis, Sparse-Dense Decoupling, Robust VQ, and Residual Integration.

\subsubsection{1. Sensitivity-Based Importance Metric}
To identify parameters critical to the loss landscape, we formulate an importance metric $\mathbf{I}$ that balances magnitude with curvature sensitivity. We normalize weights channel-wise by scale $\mathbf{s} \in \mathbb{R}^{d_{out}}$, yielding $\mathbf{W}_{norm}$. The importance score is defined as:
\begin{equation}
    \mathbf{I}_{ij} = |\mathbf{W}_{norm, ij}| \cdot \sqrt{\mathbf{H}_{ii}}
\end{equation}
This metric serves as a proxy for the change in loss $\Delta \mathcal{L}$ induced by perturbation, effectively ranking parameters by their energy contribution in the Hessian-metric space.

\subsubsection{2. Hessian-Masked Decoupling}
Let $\rho$ be the sparsity ratio. We identify the set of outlier indices $\Omega$ corresponding to the top-$\rho$ elements of $\mathbf{I}$. We decompose $\mathbf{W}_{norm}$ into a dense body $\mathbf{B}$ for quantization:
\begin{equation}
    \mathbf{B}_{ij} = 
    \begin{cases} 
    0 & \text{if } (i,j) \in \Omega \quad \text{(Masked Isolation)}\\
    \mathbf{W}_{norm, ij} & \text{if } (i,j) \notin \Omega
    \end{cases}
\end{equation}
By setting outliers to exactly zero in $\mathbf{B}$, we prevent the K-Means clustering from drifting toward extreme values, thereby preserving fidelity for the bulk distribution which contains the majority of the model's latent knowledge.

\subsubsection{3. Robust Statistical Vector Quantization}
The body $\mathbf{B}$ is blocked and clustered into codebook $\mathcal{C}$ using K-Means. To ensure rigorous convergence on limited calibration data, we employ \textbf{Stability-Bounded Sampling}, ensuring the sample size $N$ satisfies $N \gg K$ to minimize centroid variance, alongside a \textbf{Dead Unit Revival} mechanism that relocates inactive centroids to regions of maximum reconstruction error.

\subsubsection{4. Residual Sparse Feedback}
A critical innovation of HAS-VQ is the handling of outliers via residual correction. Rather than simply storing the original values, we compute the \textit{residual error} of the quantized body at the sensitive indices. Let $\mathcal{Q}(\mathbf{B})$ be the quantized reconstruction of the body. The sparse correction $\mathcal{S}$ is defined as:
\begin{equation}
    \mathcal{S}_{ij} = \mathbf{W}_{norm, ij} - \mathcal{Q}(\mathbf{B})_{ij}, \quad \forall (i,j) \in \Omega
\end{equation}
This ensures that the final reconstruction at sensitive indices is mathematically exact (up to FP16 precision): 
\begin{equation}
    \widehat{\mathbf{W}}_{ij} = \mathcal{Q}(\mathbf{B})_{ij} + \mathcal{S}_{ij} = \mathbf{W}_{norm, ij}
\end{equation}
This feedback loop effectively eliminates quantization noise in the dimensions with the highest Fisher Information.

\begin{algorithm}[h!]
\caption{HAS-VQ Compression Pipeline}
\begin{algorithmic}[1]
\State \textbf{Input:} $\mathbf{W}$, Hessian $\mathbf{H}$, sparsity $\rho$, blocks $b$, centroids $K$
\State $\mathbf{s} \gets \max(|\mathbf{W}|, \text{dim}=1)$; \quad $\mathbf{W}_{norm} \gets \mathbf{W} / \mathbf{s}$
\State $\mathbf{I} \gets |\mathbf{W}_{norm}| \odot \sqrt{\mathbf{H}}$ \Comment{Compute Sensitivity}
\State $\Omega \gets \text{TopK}(\mathbf{I}, \rho)$ \Comment{Identify sensitive indices}
\State $\mathbf{B} \gets \mathbf{W}_{norm}$; \quad $\mathbf{B}[\Omega] \gets 0$ \Comment{Masked Decoupling}
\State $\mathcal{C}, \mathcal{IDX} \gets \text{RobustKMeans}(\mathbf{B}, K, b)$
\State $\mathbf{R} \gets \mathcal{C}[\mathcal{IDX}]$ \Comment{Reconstruction}
\State $\mathcal{S} \gets (\mathbf{W}_{norm} - \mathbf{R})[\Omega]$ \Comment{Compute exact residual}
\State \Return $\mathbf{s}, \mathcal{C}, \mathcal{IDX}, (\Omega, \mathcal{S})$
\end{algorithmic}
\end{algorithm}

\section{Experiments}

We evaluate HAS-VQ on \texttt{SmolLM2-1.7B-Instruct}. We compare effective Bits-Per-Parameter (BPP)—calculated as the aggregate storage of codebooks, indices, scales, and sparse residuals divided by parameter count—against Perplexity (PPL) on \texttt{wikitext-2 \cite{merity2016pointer}}.

\subsection{Experimental Results}

Table \ref{tab:main_results} details the performance. The compression ratio (CR) is calculated relative to the 16-bit half-precision baseline ($CR = 16.0 / \text{BPP}$).

\begin{table}[h]
\centering
\caption{Results on SmolLM2-1.7B. \textbf{BPP} represents total effective bits per parameter. \textbf{Ratio} is relative to FP16. HAS-VQ exhibits two operating points: Pareto dominance over INT4 and High-Fidelity compression comparable to FP16.}
\label{tab:main_results}
\resizebox{0.95\textwidth}{!}{%
\begin{tabular}{lccccc}
\toprule
\textbf{Method} & \textbf{Configuration} & \textbf{PPL} $\downarrow$ & \textbf{BPP} $\downarrow$ & \textbf{Ratio} $\uparrow$ & \textbf{Regime} \\
\midrule
FP16 & Oracle & 10.04 & 16.00 & 1.0x & Lossless \\
INT4 & RTN Baseline & 20.03 & 4.71 & 3.4x & Lossy (Degraded) \\
\midrule
\textbf{HAS-VQ} & \textbf{Mid (Balanced)} & \textbf{14.23} & \textbf{4.23} & \textbf{3.8x} & \textbf{Efficient Dominance} \\
\textbf{HAS-VQ} & \textbf{High (Fidelity)} & \textbf{10.12} & \textbf{7.03} & \textbf{2.3x} & \textbf{Near-Lossless} \\
\bottomrule
\end{tabular}%
}
\end{table}

\subsection{Analysis}

\paragraph{Pareto Dominance over Integer Constraints} 
The \textbf{HAS-VQ (Mid)} configuration demonstrates strict Pareto dominance over the standard INT4 baseline. 
\begin{enumerate}
    \item \textbf{Accuracy:} It reduces Perplexity from 20.03 (INT4) to 14.23, recovering significant linguistic capability.
    \item \textbf{Efficiency:} It achieves this with a smaller memory footprint (4.23 bits) than the INT4 baseline (4.71 bits).
\end{enumerate}
This result confirms that the VQ manifold, when augmented with Hessian-guided residuals, provides a superior rate-distortion curve compared to the rigid integer grid inherent to standard quantization baselines.

\paragraph{High-Fidelity Compression (The 7 BPP Regime)}
Crucially, the \textbf{HAS-VQ (High)} experiment demonstrates that our method can serve as a near-lossless compression format. By increasing the bit-budget to 7.03 BPP, HAS-VQ achieves a perplexity of 10.12, which is statistically indistinguishable from the Oracle FP16 perplexity of 10.04. This represents a $\mathbf{2.3\times}$ reduction in model size without the performance penalties typically associated with quantization. This result is significant for bandwidth-limited deployment scenarios where model accuracy cannot be compromised, yet storage capacity is finite.

\section{Discussion}

\subsection{Interpretation of Results}
Our experimental results highlight the limitations of integer-based quantization for small-scale LLMs. The Gaussian-like distribution of the weight "body" is inefficiently mapped by uniform integer grids. HAS-VQ addresses this by allocating bits via density-matching (VQ) and explicitly handling high-curvature weights via the sparse residual stream. The ability to recover FP16-level performance at 7.03 BPP suggests that the intrinsic dimension of the weight distribution is significantly lower than 16 bits, provided the quantization noise is spectrally shaped to avoid high-Hessian directions.

\subsection{Architectural Generality}
The mathematical formulation of HAS-VQ relies solely on linear algebra operations—specifically the Singular Value Decomposition (implicit in Hessian approximation) and K-Means clustering—rather than Transformer-specific heuristics. Consequently, this method is theoretically applicable to any architecture dominated by linear layers, including Mixture-of-Experts (MoE) and Convolutional Neural Networks (CNNs).

\subsection{Implications for Edge Deployment}
The storage savings offered by HAS-VQ enable the deployment of 1.7B-scale models on devices with highly constrained memory budgets. The \textit{High Fidelity} configuration (7.03 BPP) is particularly relevant for scenarios requiring "download-and-run" efficiency, reducing the data transfer requirement by over 50\% compared to FP16 weights while ensuring the end-user experiences the full capability of the uncompressed model.

\section{Conclusion}
We presented HAS-VQ, a method utilizing Hessian-Masked Decoupling to achieve rigorous outlier separation. Our results on SmolLM2-1.7B are twofold: (1) We establish a new Pareto frontier relative to integer baselines, outperforming INT4 in both size (-11\%) and accuracy (+29\% PPL improvement); and (2) We demonstrate a high-fidelity operating point at 7.03 BPP that matches FP16 performance with $2.3\times$ compression. These findings suggest that HAS-VQ is a robust candidate for both aggressive compression and high-fidelity storage reduction in edge AI applications.

\bibliographystyle{plain}
\bibliography{references} 
\end{document}